\documentclass[runningheads]{llncs}
\usepackage[T1]{fontenc}
\usepackage{pifont}
\newcommand{\cmark}{\ding{51}} 
\newcommand{\xmark}{\ding{55}} 
\usepackage{graphicx}
\usepackage{amssymb}
\usepackage{graphicx}
\usepackage{caption}
\usepackage{subcaption}
\usepackage{multirow}
\usepackage{rotating}
\usepackage{float}
\usepackage{amsmath,amsfonts}
\usepackage{algorithmic}
\usepackage{algorithm}
\usepackage{array}
\usepackage[utf8]{inputenc}
\usepackage{tabularx}
\usepackage{booktabs}

\usepackage{amsmath}

\usepackage{amssymb}
\usepackage{pifont}
\usepackage{textcomp}
\usepackage{stfloats}
\usepackage{url}
\usepackage{verbatim}
\usepackage{graphicx}
\usepackage{cite}
\usepackage{xcolor}
\usepackage{soul}
\usepackage{caption} 
\usepackage{orcidlink}
\begin{document}

\title{Sphere-Depth: A Benchmark for Depth Estimation Methods with Varying Spherical Camera Orientations
}
\titlerunning{Sphere-Depth}

\author{Soulayma Gazzeh\orcidID{ 0000-0002-2066-669X} \and 
Giuseppe Mazzola\orcidID{0000-0003-3839-9312} \and
Liliana Lo Presti\orcidID{0000-0003-0833-4403}\and
Marco La Cascia\orcidID{0000-0002-8766-6395}}
\institute{Department of Engineering, University of Palermo, Palermo, Italy\\ \email{soulayma.gazzeh@unipa.it; liliana.lopresti@unipa.it}}
\maketitle              

\begin{abstract}
Reliable depth estimation from spherical images is crucial for 360° vision in robotic navigation and immersive scene understanding. However, the onboard spherical camera can experience unintentional pose variations in real-world robotic platforms that, along with the geometric distortions inherent in equirectangular projections, significantly impact the effectiveness of depth estimation.

To study this issue, a novel public benchmark, called Sphere-Depth, is introduced to systematically evaluate the robustness of monocular depth estimation models from equirectangular images in a reproducible way. 
Camera pose perturbations are simulated and used to assess the performance of a popular perspective-based model, Depth Anything, and of spherical-aware models such as Depth Anywhere, ACDNet, Bifuse++, and SliceNet. Furthermore, to ensure meaningful evaluation across models, a depth calibration-based error protocol is proposed to convert predicted relative depth values into metric depth values using supervised learned scaling factors for each model.

Experiments show that even models explicitly designed to process spherical images exhibit substantial performance degradation when variations in the camera pose are observed with respect to the canonical pose. 
The full benchmark, evaluation protocol, and dataset splits are made publicly available at: \url{https://github.com/sgazzeh/Sphere_depth}.
\end{abstract}

\section{Introduction}

\begin{figure}[ht]
\centering
\setlength{\tabcolsep}{3pt} 
\renewcommand{\arraystretch}{0.0}
\begin{minipage}{0.3\textwidth}
    \textbf{(a) Planar Image}\\[1ex]
    \centering
    \includegraphics[height=0.48\textwidth]{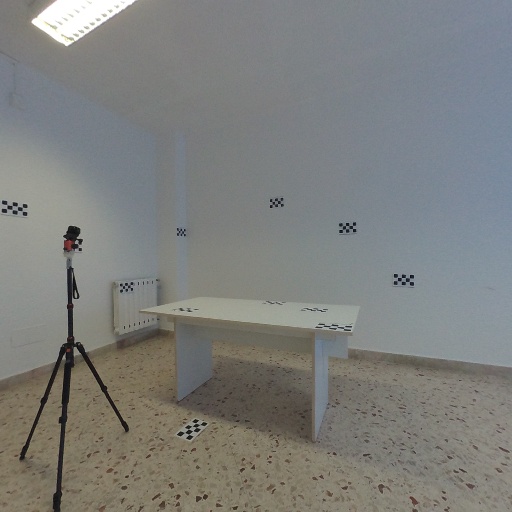} 
\end{minipage}
\begin{minipage}{0.3\textwidth}
    \textbf{(b) Canonical Pose} \\[1ex]
    \centering
    \includegraphics[width=\textwidth]{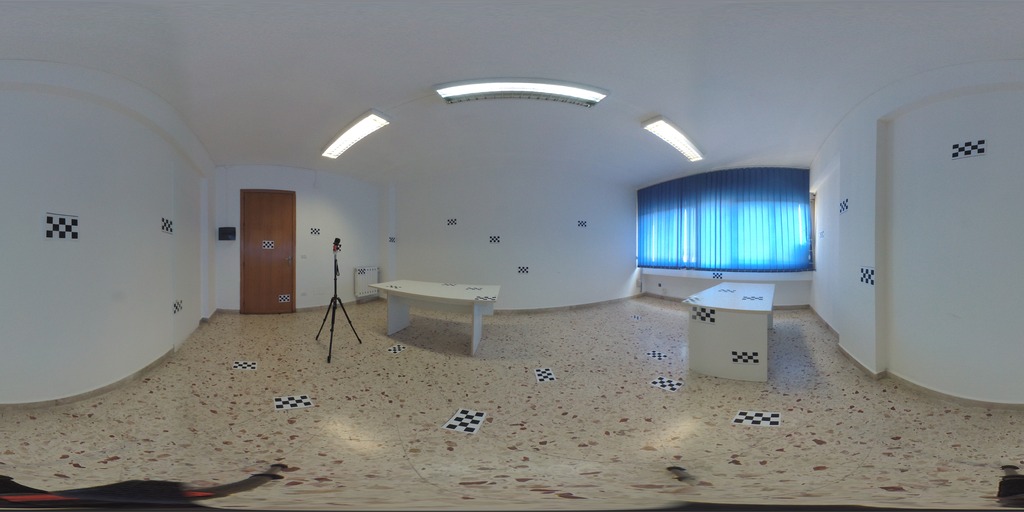} 
\end{minipage}
\begin{minipage}{0.3\textwidth}
    \textbf{(c) Changing Pose}\\[1ex]
    \centering
    \includegraphics[width=\textwidth]{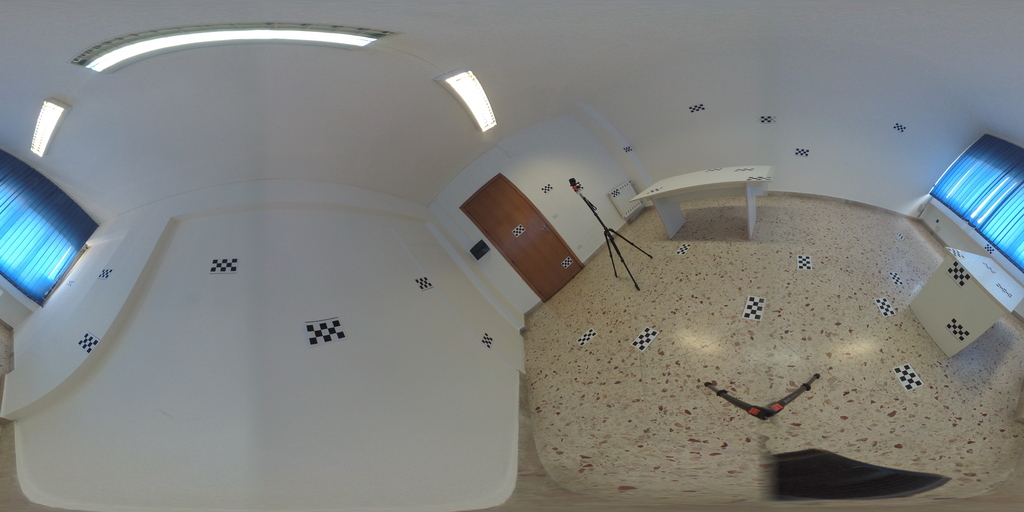} 
\end{minipage}
\caption{(a) planar image obtained by cubic projection; (b) gravity-aligned image obtained with a canonical camera pose; (c) Equirectangular image with variations in camera pose.}
\label{fig:intro}
\end{figure}

Depth estimation from spherical images has emerged as a key component to achieve immersive 360° scene understanding and autonomous navigation~\cite{wang2024depth, ai2024elite360d, mohadikar2025omnidiffusion}. 
Unlike conventional pinhole cameras, which provide a limited field of view and require multiple frames to capture the complete scene, spherical cameras provide continuous and full scene coverage.
This capability is particularly valuable for robotic platforms operating in dynamic, unconstrained environments, such as autonomous vehicles~\cite{zheng2024physical}, drones~\cite{jaisawal2024airfisheye}, and mobile robots. 

The ability to accurately estimate depth values from spherical images is useful in tasks such as scene understanding, object detection~\cite{park2024odd}, and motion planning~\cite{cho2024obstacle}, all of which are foundational to successful autonomous operation.

Current depth estimation models~\cite{yang2024depth,wang2025pddepth, patni2024ecodepth, ai2023hrdfuse, wang2022bifuse++, zhuang2022acdnet} operate under distinct assumptions about the camera geometry. 
{\em Perspective-based models}~\cite{eigen2014depth, laina2016deeper} are typically trained on planar images.
 
Techniques such as cubemap projection ~\cite{cheng2018cube} or planar conversion ~\cite{su2016pano2vid} are employed to flatten the spherical view into planar images (Fig.~\ref{fig:intro}.a) that can be processed by these models.
This is needed because the equirectangular projection introduces geometric distortions in images that significantly alter spatial relationships in the scene and affect depth prediction accuracy, especially near the poles.

Recent {\em Spherical-aware models} extend perspective-based models to handle equirectangular images by adopting distortion-aware sampling~\cite{athwale2023darswin} or tailored data augmentation techniques~\cite{wang2024depth}, by projecting features onto tangent planes to locally approximate perspective projection~\cite{eder2020tangent}, by dividing the equirectangular image into vertical slices~\cite{pintore2021slicenet} or adopting specific padding techniques~\cite{zhuang2022acdnet}, by modifying the convolutional operation to consider the spherical geometry~\cite{cohen2018spherical} or by using latitude-dependent~\cite{yoon2022spheresr} or rectangular-shaped kernels~\cite{zioulis2018omnidepth}. Additionally, methods~\cite{wang2022bifuse++,jiang2021unifuse} merge features extracted from cubemap and equirectangular projections, while models~\cite{li2022omnifusion,ai2023hrdfuse} process image patches, and use geometric priors to merge the patch-based depth maps into a final depth map.

The previous approaches mostly assume that spherical cameras acquire equirectangular images in a canonical pose, i.e., with the equatorial plane parallel to the floor plane. Images acquired under this assumption are called {\em gravity-aligned images} (Fig.~\ref{fig:intro}.b).
However, this assumption may not hold in practice. For instance, spherical cameras mounted on mobile platforms can be subjected to uncontrollable pose perturbations (namely, camera orientation variations) due to the platform's motion. These perturbations significantly change the appearance of the acquired equirectangular images (Fig.~\ref{fig:intro}.c), and compromise the accuracy of the predicted depth values. The issue has received little attention in the literature, and prevents a full understanding of how well depth estimation models perform in practice.

In this paper, we introduce a novel benchmark designed to systematically evaluate the robustness of popular depth estimation models (especially spherical-aware models) under a range of camera pose variations in real settings. In contrast to existing benchmarks for 360° depth estimation, such as \cite{albanis2021pano3d}, our framework introduces structured pitch and roll rotations to simulate different levels of camera pose variations and measure errors in realistic settings by using known 3D point coordinates of landmarks displaced in the scene.

In this paper, our main contributions are:
\begin{itemize}
    \item \textbf{A pose-aware benchmark} for monocular depth estimation from equirectangular images, designed to evaluate model generalization under camera pose variation with a focus on sensitivity to pitch and roll changes.
    
    \item \textbf{A depth calibration-based evaluation protocol} that introduces a learnable scaling factor $\lambda$ to align predicted depths to the ground truth values, allowing fair comparison between models with different output scales.
    \item \textbf{A comparative study} of models trained on either perspective or spherical images concluding that even the performance of models designed to process spherical images degrades significantly in the presence of camera pose perturbations.

\end{itemize}

Our results highlight that despite recent advances, current 360° depth estimation models still implicitly rely on cameras in a canonical pose and are not robust to spatial distortions caused by pose variations. This public benchmark provides a reproducible basis for future research on pose- and geometry-invariant depth estimation in omnidirectional settings.

\section{Depth Estimation Models selected for benchmarking}
To assess the state-of-the-art in depth estimation from equirectangular images, we selected a representative set of spherical-aware models based on their architectural diversity, input modalities, and strategies to handle spherical distortions, as summarized in Table~\ref{tab:model_comparison}. 

We also consider a popular perspective-based model, Depth Anything v2~\cite{yang2024depth}, to evaluate its generalization capabilities to 360° imagery via cubemap projections.

All selected methods have an encoder-decoder architecture. In Depth Anything v2, the encoder is a Vision Transformer (ViT) and the decoder is a Dense Prediction Transformer (DPT) devoted to the regression of depth values. The model does not explicitly handle equirectangular projection (ERP).

BiFuse++~\cite{wang2022bifuse++} is a two-branch encoder-decoder that geometrically aligns features extracted from the equirectangular image and features from the cube face images, while DepthAnywhere~\cite{wang2024depth} exploits a transformer-based model within a teacher-student framework. The model processes equirectangular images. At training time, it uses Depth Anything as a teacher to independently process each of the cube faces and generate pseudo-labels. It also adopts data augmentation by randomly rotating the spherical images.

Other methods that process only equirectangular images are
ACDNet~\cite{zhuang2022acdnet} and SliceNet~\cite{pintore2021slicenet}. They both adopt a ResNet-based encoder. In ACDNet, the decoder exploits Adaptively Combined Dilated Convolutional layers (ACDConv) to obtain adaptive receptive fields. Furthermore, circular padding of the images is used to deal with the image circularity introduced by ERP. SliceNet includes multi-scale vertical feature slicing and a ConvLSTM-based decoder. The model directly processes gravity-aligned images. 
\begin{table}[t!]
\centering
\caption{Theoretical comparison of selected depth estimation models.}
\label{tab:model_comparison}
\resizebox{\textwidth}{!}{
\begin{tabular}{|l|c|p{4.5cm}|p{5cm}|}
\hline
\textbf{Model} & \textbf{Full 360° Image?} &\textbf{Architecture} & \textbf{Handling Spherical Geometry} \\
\hline
\hline
\textbf{Depth Anything v2 \cite{yang2024depth}} & \xmark & ViT + DPT & No ERP support; trained on perspective images. \\
\hline
\textbf{DepthAnywhere \cite{wang2024depth}} & \cmark & Transformer-based teacher-student framework using~\cite{yang2024depth} as teacher on unlabeled data & Depth Anything on unlabeled cube face images; random rotations of spherical images. \\
\hline
\textbf{BiFuse++ \cite{wang2022bifuse++}} & \cmark & Two-branch encoder + decoder (ResNet-34)
& Process both equirectangular and cube face images; geometrical align features from different input \\
\hline
\textbf{ACDNet \cite{zhuang2022acdnet}} & \cmark & ResNet + ACDConv-based decoder & ACDConv and circular padding mitigate ERP distortions. \\
\hline
\textbf{SliceNet \cite{pintore2021slicenet}} & \cmark & ResNet-based encoder + multi-scale vertical feature slicing + ConvLSTM-based decoder. & Leverages gravity-aligned slices without explicit projections. \\
\hline
\end{tabular}
}
\end{table}

\subsection{Depth Value Extraction strategies}
The evaluated models differ in how they represent and extract depth information. They either learn relative disparities or generate direct metric depth maps. To maintain consistency, we adhere to the recommended protocols for each model. 

\textbf{Direct Depth Prediction:} 
The models~\cite{pintore2021slicenet, zhuang2022acdnet} generate metric depth maps directly from their respective architectures.

\textbf{Disparity-Based Estimation:} As done in these papers ~\cite{yang2024depth,wang2024depth,wang2022bifuse++}, the depth is estimated using intermediate disparity representations. The raw outputs $\hat{d}$ are converted to depth
value d by the following inverse transformation:
\begin{equation*}
   d = \frac{1}{\alpha\sigma(\hat{d}) + \epsilon}
\end{equation*}
with $\sigma(\cdot)$ the sigmoid activation function, $\alpha$ a model-specific scaling factor, and $\epsilon$ a small constant added for numerical stability. All predicted values are clipped to the range $[0, 10]$ meters. 

\section{Dataset, Depth Calibration and Experimental Results}
The dataset includes 8 high-resolution equirectangular images acquired indoors with two Garmin VIRB 360 cameras mounted in various locations and with different orientations.
37 landmarks located in the scene were chosen as reference points, and their 3D coordinates were measured with respect to a World coordinate system. 
For each image, the dataset includes manually selected pixel coordinates of visible landmarks, the 3D position of the camera, and its orientation (yaw, pitch, roll) relative to the chosen World coordinate system. Thus, it is possible to determine the distance (depth) of each landmark to the camera. 

To address scale ambiguity in monocular depth estimation, we estimate a scaling factor $\lambda$ that aligns predicted depth maps $d$ with our ground-truth depth $d_{\text{gt}}$. To this purpose, we split the landmarks visible in the image in a training and a test set. The training landmarks are used to estimate $\lambda$ by minimizing the mean squared error in the least squares sense. $\lambda$ is defined as:

\begin{equation*}
    \lambda = \frac{\sum_{i}^M d_i  d_{\text{gt},i}}{\sum_{i}^M d_i^2}
\end{equation*}
where $i$ indexes the M available training landmarks.
It is important to note that $\lambda$ is computed using gravity-aligned images. 
Gravity-aligned images are computed by aligning the camera's reference axes with the 3D World coordinate system through a 3D rotation of the spherical image.

For each image, we measured the mean squared error between the predicted scaled depth value $\lambda d$ and the true depth value $d_{gt}$ over the test landmarks. In our experiments, we report the average error value $\varepsilon$ over the processed images.
\begin{table}[b!]
\footnotesize
\caption{Depth error $\varepsilon$ (meters) on gravity-aligned and real images}
\label{tab:deformation}
\centering
\begin{tabular}{l|c|c|c|c}
\toprule
 & $\lambda$ & $\varepsilon$(G. Aligned)&$\varepsilon$(Small Def.)&$\varepsilon$(High Def.) \\
\midrule
\textbf{ACDNet}~\cite{zhuang2022acdnet} & 1.11 & \textbf{0.21} & \textbf{0.23} & \textbf{1.38} \\
\textbf{DepthAnywhere}~\cite{wang2024depth} & 0.56 & 0.26 & 0.34 & 2.32 \\
\textbf{Bifuse++}~\cite{wang2022bifuse++} & 0.93 & 0.38 & 0.32 & 2.04 \\
\textbf{SliceNet}~\cite{pintore2021slicenet} & 0.92 & 0.41 & 0.65 & 2.09 \\
\bottomrule
\end{tabular}%
\end{table}

\subsection{Depth Estimation from equirectangular images}
We evaluate the robustness of spherical-aware depth estimation models on the 8 real images from our dataset and corresponding gravity-aligned images. Real images were split based on the degree of camera orientation variation into 4 {\em Small deformation} images (with pitch and roll angles in the range ±1°) and 4 {\em High deformation} images (with pitch and roll angles exceeding ±15°, up to ±30°).

For each model, Table~\ref{tab:deformation} presents the learned scaling depth factor $\lambda$, and the error $\varepsilon$ measured in the above described scenarios. 
For all models except DepthAnywhere, the $\lambda$ value is close to 1, indicating a good alignment of the ranges of the predicted depths with the true value range. DepthAnywhere consistently underestimates depth being $\lambda=0.56$. 

ACDNet performs well on gravity-aligned images, with a measured error of $\varepsilon=0.21$. Its accuracy decreases as the camera orientation changes. However, on our dataset, ACDNet outperforms all other selected techniques.
Indeed, DepthAnywhere, BiFuse++, and SliceNet struggle with large camera orientation variations, while achieving comparable errors on gravity-aligned images. Among all the models, SliceNet struggles even with small camera pose changes.

\begin{figure*}[t!]
    \centering
    \includegraphics[height=0.25\linewidth]{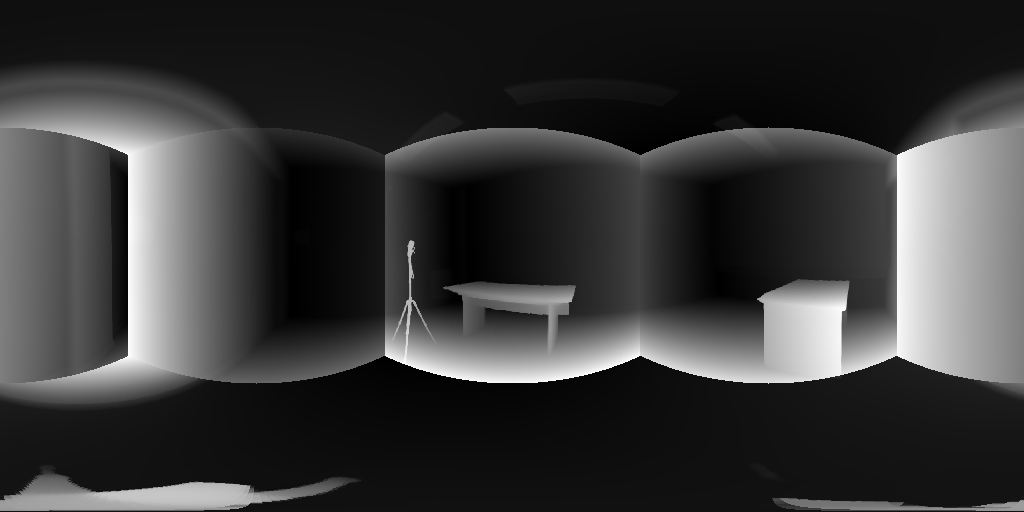}    \caption{Equirectangular depth map of the scene obtained by reprojecting six depth maps of the cube faces. }
    \vspace{-10pt}
    \label{6facedepth}
\end{figure*} 
\subsection{Depth Estimation from cube face images}
In this experiment, we project a gravity-aligned 360° image into six cube faces and apply the \textit{Depth Anything} model to generate depth predictions for each face. 

These predicted cubic depth maps are subsequently re-projected into an equirectangular format to reconstruct the spherical depth map of the scene (Fig.~\ref{6facedepth}). As in the previous experiment, the error is calculated on the test landmarks.
In this case, the estimated scale factor is $\lambda=1.38$, which means that, in general, Depth Anything tends to underestimate depth values, probably because it cannot use contextual information about the relationships between the cube faces. The estimated error $\varepsilon$(G. Aligned)$=1.72$ and is understandably high considering the erroneously estimated depth values along the face borders.
These findings reveal the limitations of perspective models with spherical data, as independently processing cube faces fails to account for the scene's geometry. 

\subsection{Depth Estimation under Spherical Camera Pose Variation}
To further investigate the robustness of depth estimation models against camera orientation changes, we conducted experiments simulating pitch and roll variations. Controlled rotations were applied to 360° equirectangular images to mimic camera misalignment with the gravity vector. The pitch and roll angles were varied in two sets:

(1) High changes from -40° to +40° in 10° increments.

(2) Small changes from -2° to +2° in 0.5° increments.

This experimental study examines each model's sensitivity to both significant and subtle orientation changes encountered in real-world scenarios using a single image sample.
\begin{figure}[t!]
\centering
\setlength{\tabcolsep}{1pt}
\renewcommand{\arraystretch}{0.0}

\begin{minipage}{\textwidth}   
    \hspace{-3.5cm}\textbf{(a) ACDNet}\\[1ex]
    \centering
    \begin{tabular}{cc}
        \includegraphics[width=0.48\textwidth]{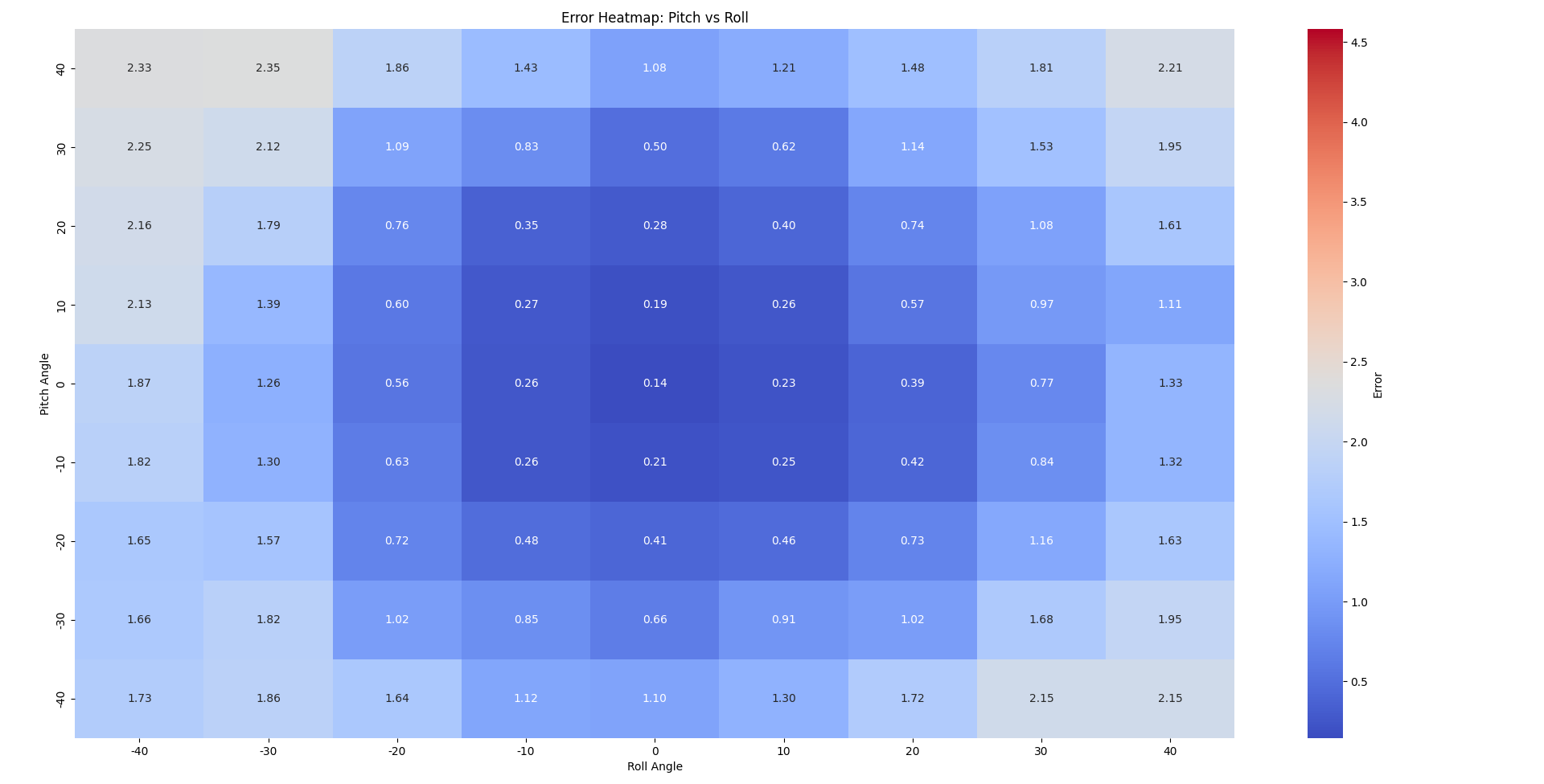} &
        \includegraphics[width=0.38\textwidth]{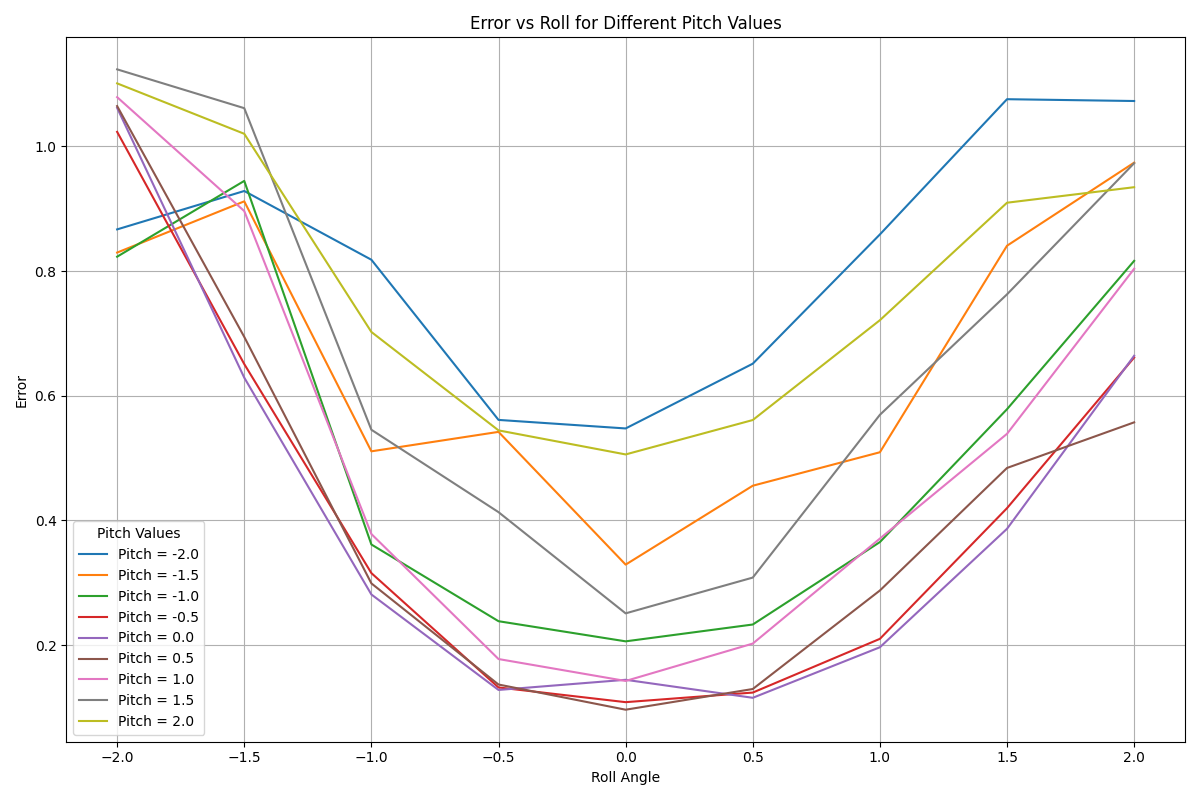}
    \end{tabular}
\end{minipage}

\vspace{0.5cm}

\begin{minipage}{\textwidth}
    \hspace{-3.5cm}\textbf{(b) SliceNet}\\[1ex]
    \centering
    \begin{tabular}{cc}
        \includegraphics[width=0.48\textwidth]{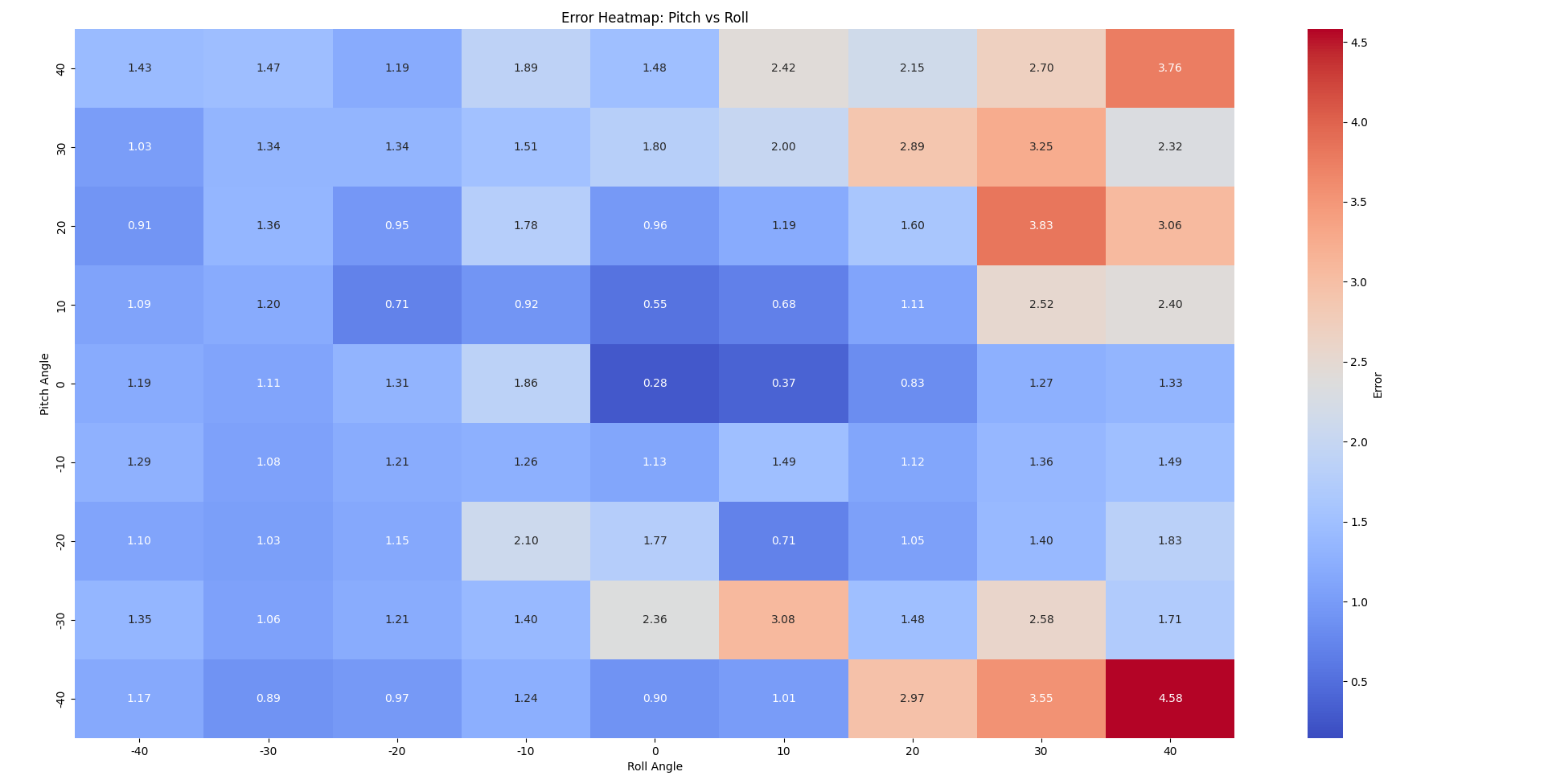} &
        \includegraphics[width=0.38\textwidth]{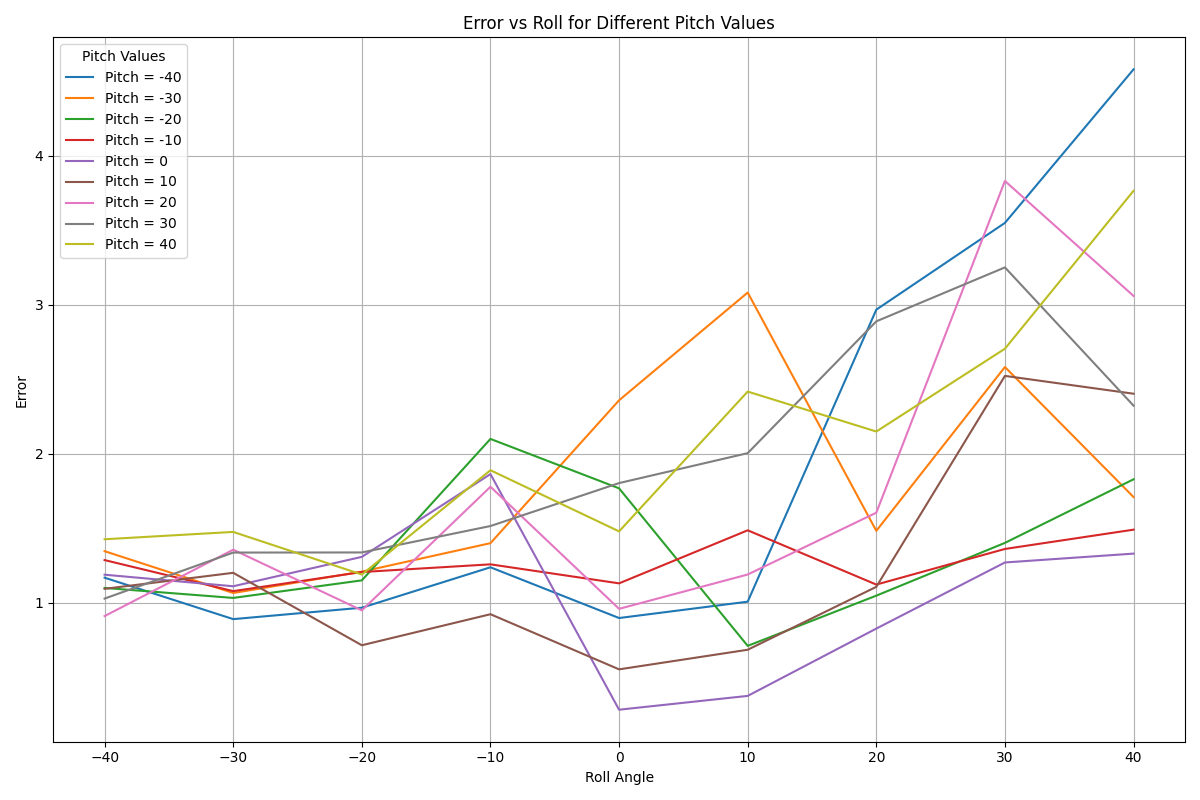}
    \end{tabular}
\end{minipage}
\caption{Error heatmaps (on the left) and error response curves (on the right) for ACDNet and SliceNet under large pitch and roll angle variations.}
\label{fig:large_variation}
\end{figure}

\subsubsection{Sensitivity to Large Camera Pose Variations}
\begin{figure}[t!]
\centering
\setlength{\tabcolsep}{2pt} 
\renewcommand{\arraystretch}{0.0}
\begin{minipage}{\textwidth}
    \hspace{-4.5cm} \textbf{(a) Canonical camera pose} \\[0.5ex]
    \centering
    \begin{tabular}{cc}
        \includegraphics[width=0.38\textwidth]{R0_P0_1.jpg} &
        \includegraphics[width=0.38\textwidth]{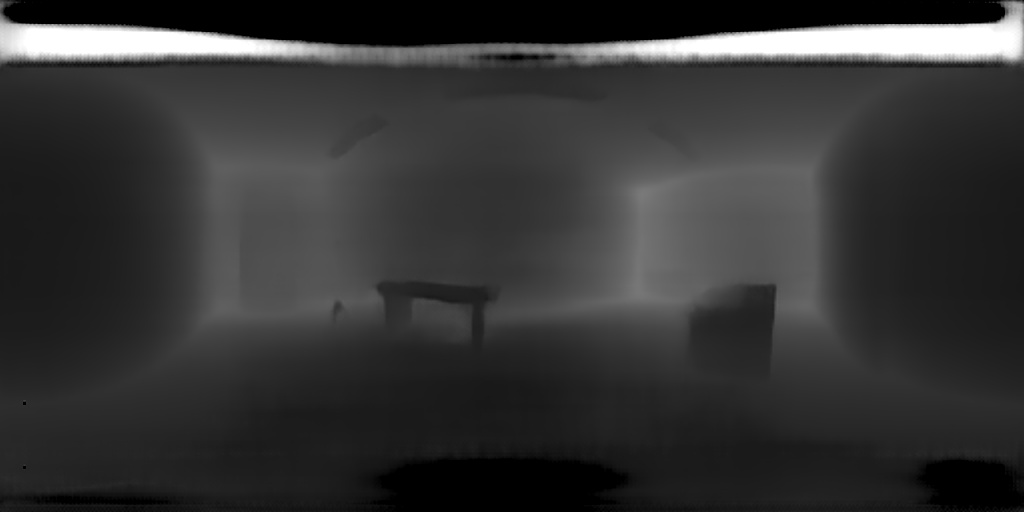}
    \end{tabular}
\end{minipage}
\vspace{0.3cm}
\begin{minipage}{\textwidth}
    \hspace{-4cm}\textbf{(b) Large camera pose variation}\\[0.5ex]
    \centering
    \begin{tabular}{cc}
        \includegraphics[width=0.38\textwidth]{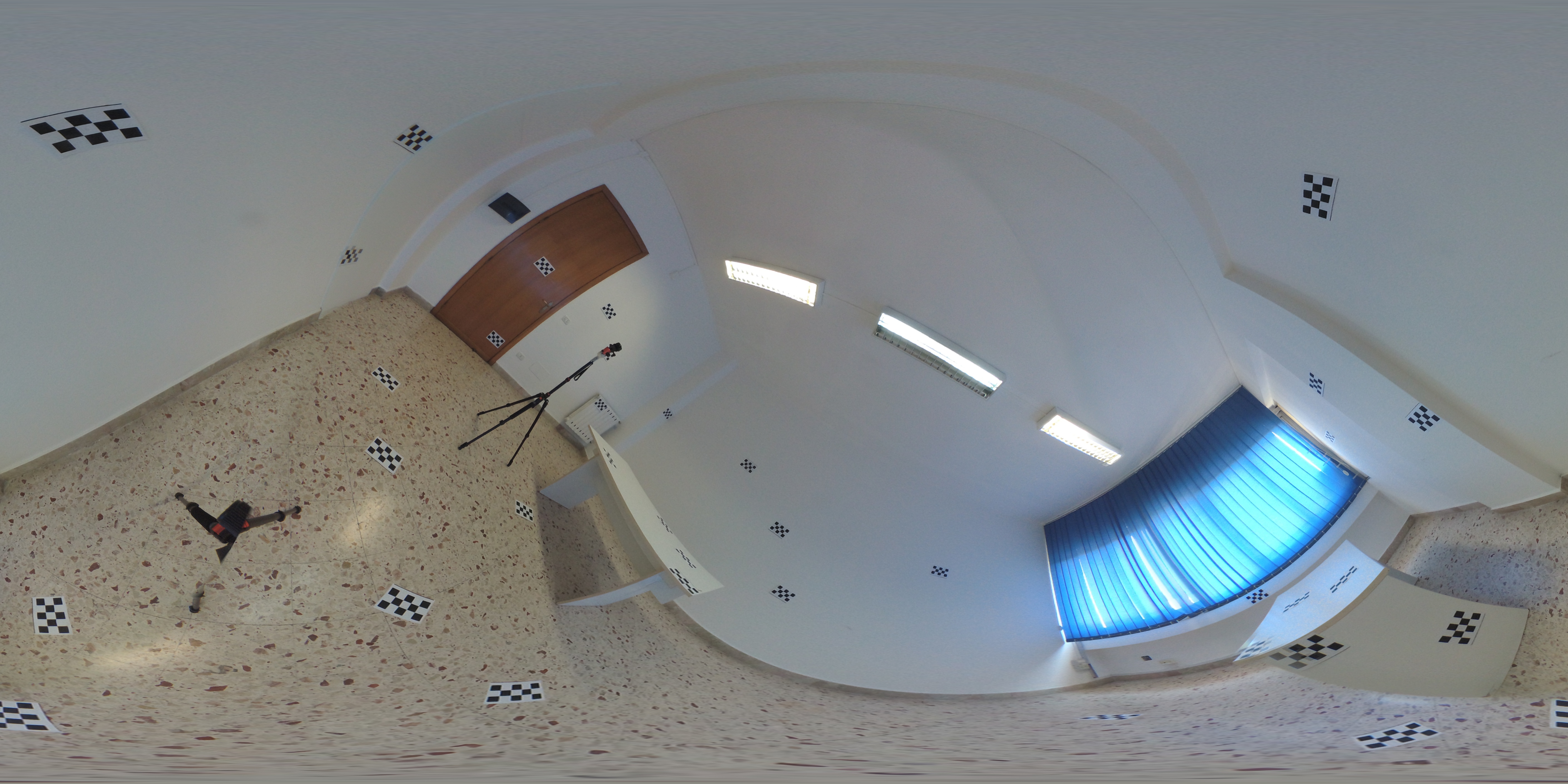} &
        \includegraphics[width=0.38\textwidth]{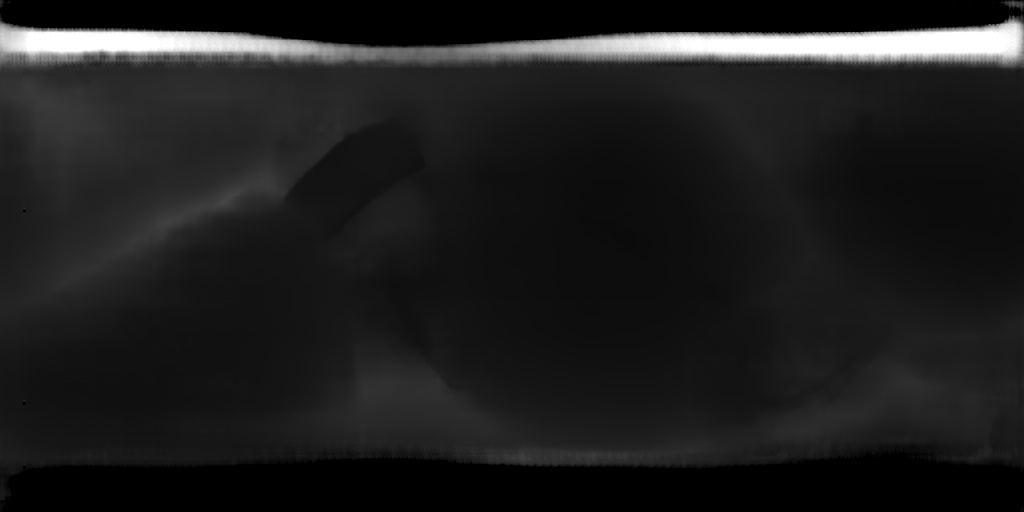}
    \end{tabular}
\end{minipage}
\caption{Qualitative results for ACDNet under conditions of no camera pose variation and significant camera pose variation.}
\label{fig:Qual}
\end{figure}

Due to the strong behavior alignment among the models' results, we present ACDNet's result as a representative example in Figure \ref{fig:large_variation}(a).
 ACDNet, DepthAnyWhere, and BiFuse++ show consistent patterns, with optimal performance near the canonical camera orientation (pitch = 0°, roll = 0°) and increasing error at the extremes, indicating sensitivity to significant pose variations while managing moderate variation well. In contrast, SliceNet behaves differently (Figure \ref{fig:large_variation}(b)), exhibiting asymmetric patterns with cooler regions in high-variation corners.
It maintains consistent performance despite considerable changes in pitch and roll and displays a monotonic increase in error with roll at positive pitch values, deviating from the typical symmetric U-shape. This suggests that SliceNet might be less sensitive to variations, but could be less robust in certain angular configurations. For completeness, qualitative results for ACDNet under no and large camera pose variation (pitch = -40°, roll = 40°) are provided in Figure \ref{fig:Qual}.

\begin{figure}[t!]
\centering
\setlength{\tabcolsep}{1pt}
\renewcommand{\arraystretch}{0.0}

\begin{minipage}{\textwidth}   
    \hspace{-3.5cm}\textbf{(a) ACDNet}\\[1ex]
    \centering
    \begin{tabular}{cc}
        \includegraphics[width=0.48\textwidth]{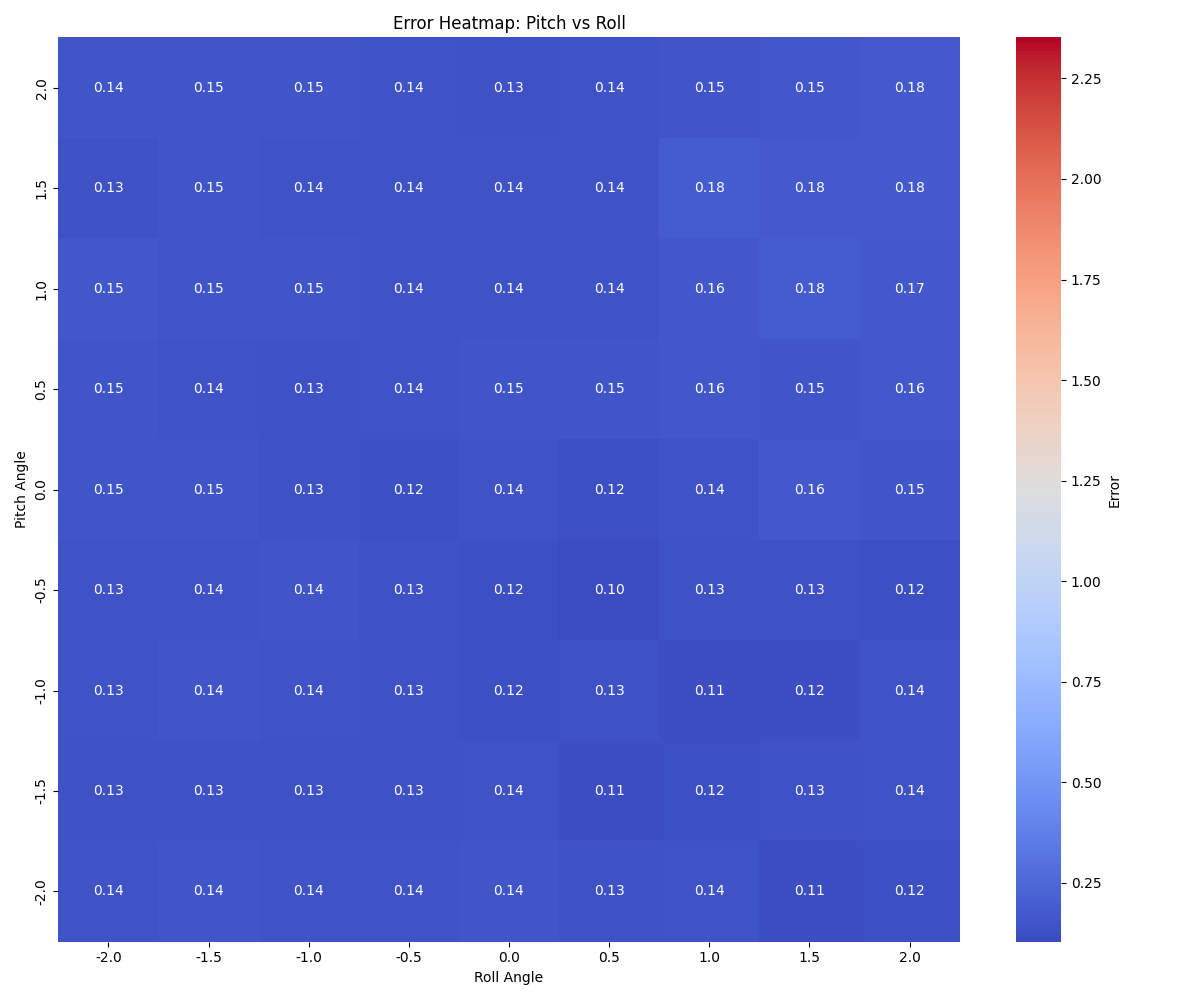} &
        \includegraphics[width=0.58\textwidth]{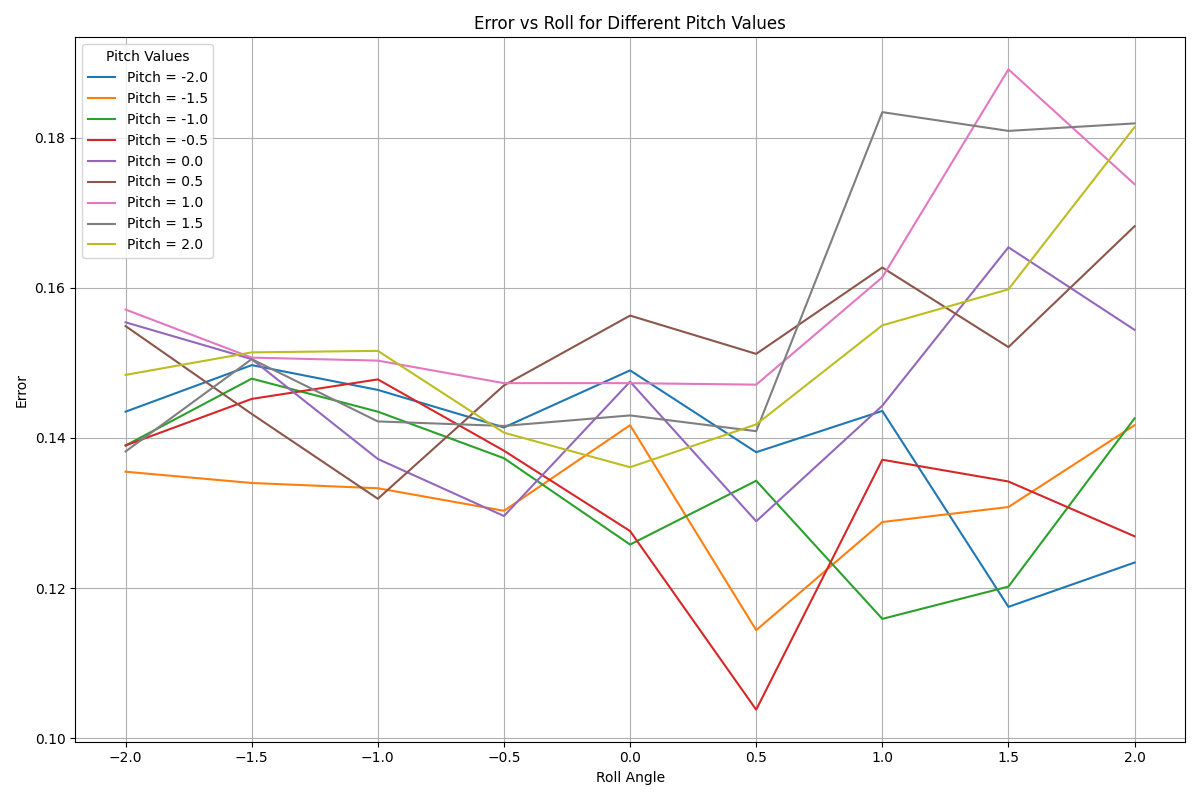}
    \end{tabular}
\end{minipage}

\vspace{0.5cm}

\begin{minipage}{\textwidth}
    \hspace{-3.5cm}\textbf{(b) Depth Anything}\\[1ex]
    \centering
    \begin{tabular}{cc}
        \includegraphics[width=0.48\textwidth]{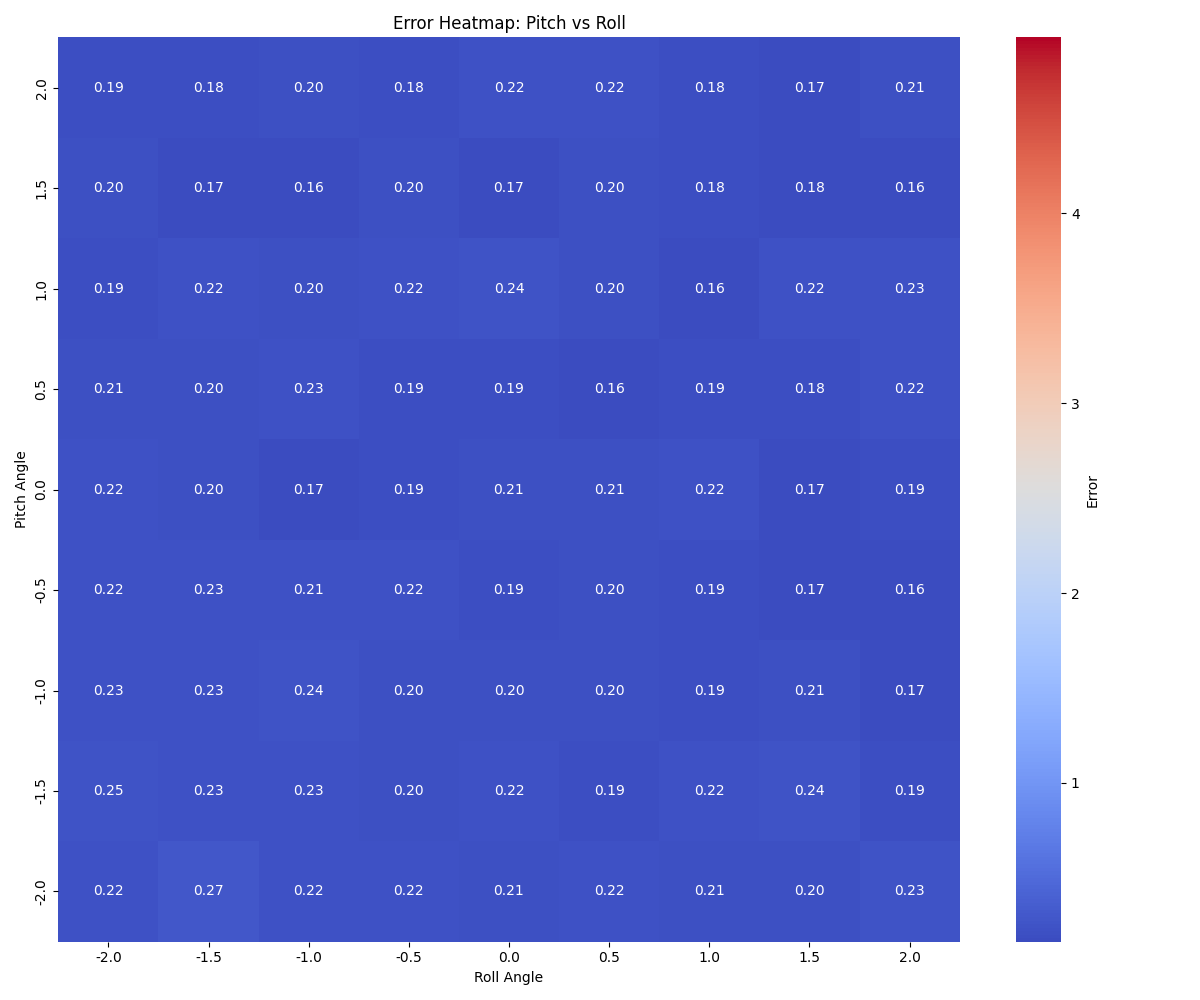} &
        \includegraphics[width=0.58\textwidth]{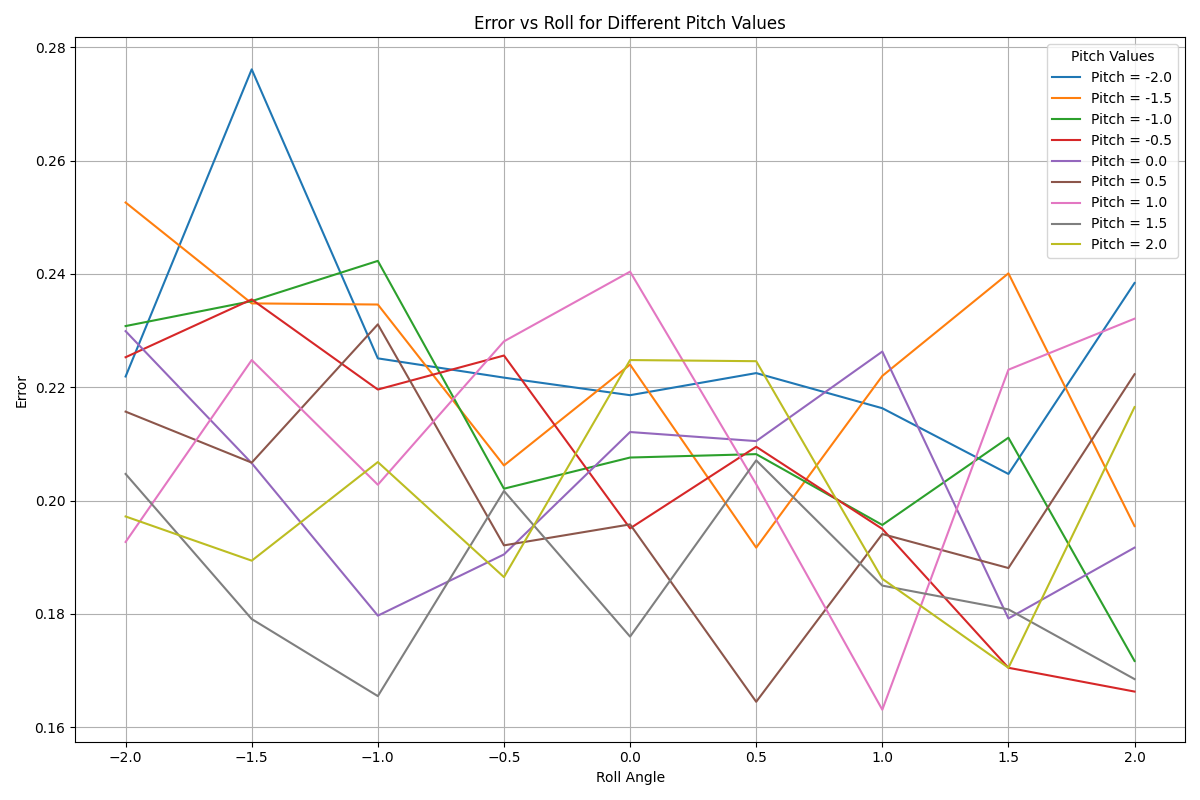}
    \end{tabular}
\end{minipage}
\caption{Error heatmaps (on the left) and error response curves (on the right) for ACDNet and Depth Anything under small pitch and roll angle variations.}
\label{fig:small_variation}
\end{figure}

\subsubsection{Sensitivity to Small Camera Pose Variations}

To evaluate depth estimation stability under small camera orientation changes, we simulated ±2° pitch and roll variations. 
While most models maintained consistent performance, ACDNet, which previously achieved the best results on gravity-aligned image with an error of $\varepsilon$= 0.14, showed a sensitivity to these perturbations.
However, we noted a relative increase of about 9–19\% when compared to the aligned baseline. 
This reflects a small but measurable degradation in performance.

Yaw variations were excluded from the experiment as they primarily cause a horizontal shift in the equirectangular image, having no significant impact on depth estimation.

\subsection{Performance Across Depth Ranges}
The LOWESS curves were computed for all data points per model to ensure a fair trend estimation, crucial for understanding depth-specific biases. This analysis indicates that while some models perform well for close-range depth recovery, others excel in distant scenes, which is vital for applications like navigation in large indoor spaces.

\begin{figure*}[t!]
    \centering
    \includegraphics[width=0.7\linewidth]{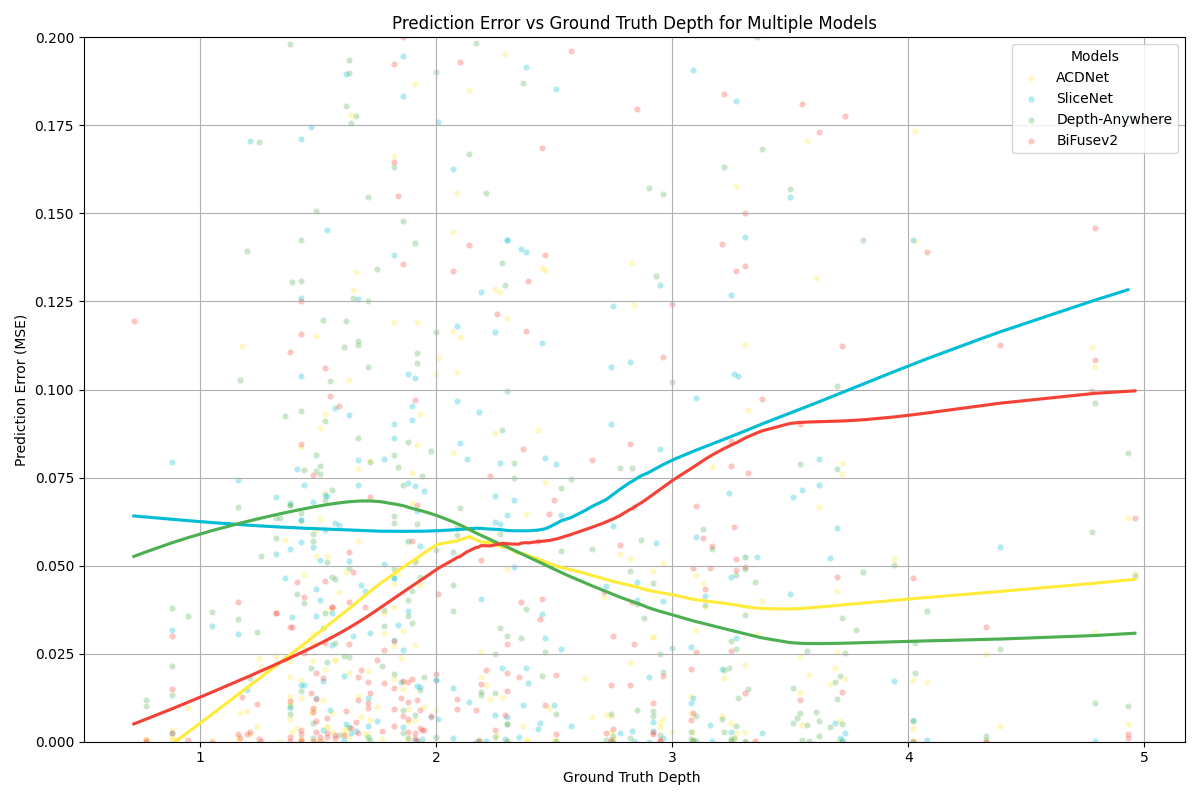}
    \caption{LOWESS curves describing error trend across equirectangular models.}
    \label{all}
\end{figure*}

We analyzed prediction errors across varying ground truth depth values to assess the models' spatial performance consistency. The LOWESS curves in Figure~\ref{all} reveal error trends relative to depth, showing which models are better for nearby or distant structures.
ACDNet and Depth-Anywhere maintain stable errors across the depth spectrum, with Depth-Anywhere slightly better in far-range regions (depth > 3m). SliceNet is effective for close-range (depth < 2m) but struggles with long-range predictions. BiFuseV2 shows increasing error, particularly beyond 2.5m, indicating less robustness in deep scenes.

\section{Conclusion}

We introduced a novel benchmark to evaluate the robustness of monocular depth estimation models on spherical imagery under realistic camera pose variations. By simulating deviations in pitch and roll, the sensitivity of models across a range of perturbation intensities is studied, capturing performance degradation trends from minor to significant orientation changes. 

Our depth calibration-aware evaluation protocol utilizes a learnable scale factor to consistently compare models with varying output magnitudes. A multi-axis error analysis reveals that depth accuracy deteriorates with pose variation and at different depth ranges, with many models biased toward canonical viewpoints.

Our results show that even advanced spherical geometry-aware models struggle to generalize beyond their training data and are vulnerable to camera pose distortions. This benchmark creates a practical testing environment, highlighting current limitations and guiding the development of pose-invariant, distortion-aware depth estimation techniques for 360° vision systems.

\section{Acknowledgments}
This work was partially supported by European Union – “Next
Generation EU” - PNRR Mission 4 “Istruzione e Ricerca” Component 2 “Dalla Ricerca all’Impresa” - Investment line 1.3 Project “Future Artificial Intelligence - FAIR”
cod. PE0000013, CUP J53C22003010006 Title: “CAESAR: Self-conscious behaviour in Embodied AI Agents.
\bibliographystyle{splncs04}
\bibliography{ref}

\end{document}